\pdfoutput=1
\documentclass[11pt]{article}

\usepackage{acl}

\usepackage{times}
\usepackage{latexsym}

\usepackage[T1]{fontenc}
\usepackage[utf8]{inputenc}

\usepackage{microtype}

\usepackage{graphicx}
\usepackage{amsmath}
\usepackage{amsthm}
\usepackage{booktabs}
\usepackage{algorithm}
\usepackage{algorithmic}
\usepackage{amsfonts,amssymb}
\usepackage{multirow}
\usepackage[utf8]{inputenc}
\usepackage{cleveref}
\crefname{section}{§}{§§}
\Crefname{section}{§}{§§}

\usepackage{amsthm,amsmath,amssymb}
\usepackage{mathrsfs}

\title{CAKE: A Scalable Commonsense-Aware Framework For Multi-View Knowledge Graph Completion}

\author{
Guanglin Niu\textsuperscript{\rm 1,2},
Bo Li\textsuperscript{\rm 2,3},
Yongfei Zhang\textsuperscript{\rm 1},
Shiliang Pu\textsuperscript{\rm 4}
\\ % All authors must be in the same font size and format. Use \Large and \textbf to achieve this result when breaking a line
\textsuperscript{\rm 1} Beijing Key Laboratory of Digital Media, Beihang University, Beijing, China\\
\textsuperscript{\rm 2} Institute of Artificial Intelligence, Beihang University, Beijing, China\\
\textsuperscript{\rm 3} Hangzhou Innovation Institute, Beihang University, Hangzhou, China\\
\textsuperscript{\rm 4}Hikvision Research Institute, Hangzhou, China\\
\{beihangngl, boli, yfzhang\}@buaa.edu.cn, pushiliang.hri@hikvision.com
}

\begin{document}
\maketitle
\begin{abstract}
Knowledge graphs store a large number of factual triples while they are still incomplete, inevitably. The previous knowledge graph completion (KGC) models predict missing links between entities merely relying on fact-view data, ignoring the valuable commonsense knowledge. The previous knowledge graph embedding (KGE) techniques suffer from invalid negative sampling and the uncertainty of fact-view link prediction, limiting KGC's performance. To address the above challenges, we propose a novel and scalable \textbf{\underline{C}ommonsense-\underline{A}ware \underline{K}nowledge \underline{E}mbedding} (\textbf{CAKE}) framework to automatically extract commonsense from factual triples with entity concepts. The generated commonsense augments effective self-supervision to facilitate both high-quality negative sampling (NS) and joint commonsense and fact-view link prediction. Experimental results on the KGC task demonstrate that assembling our framework could enhance the performance of the original KGE models, and the proposed commonsense-aware NS module is superior to other NS techniques. Besides, our proposed framework could be easily adaptive to various KGE models and explain the predicted results.
\end{abstract}

\section{Introduction}

In recent years, knowledge graphs (KGs) such as Freebase~\cite{BGF:Freebase}, DBpedia~\cite{Lehmann:dbpedia} and NELL~\cite{Mitchell:nell} have been widely used in many knowledge-intensive applications, including question answering~\cite{SPARQA, EmbedKGQA}, dialogue systems~\cite{GraphDialog, Zhou:dialogue} and recommender systems~\cite{KGIN2020, Wang:RS}. However, the KGs constructed manually or automatically are inevitably incomplete, requiring KGC to infer new facts.

The previous KGC models can be classified into three main streams: (1) Rule learning-based models mine logic rules for induction reasoning, such as AMIE+~\cite{Galarrage:AMIE}, DRUM~\cite{DRUM} and AnyBurl~\cite{AnyBurl}. (2) Path-based models~\cite{AttnPRA, DeepPath, Lin:reward-shaping} search paths for multi-hop reasoning. (3) KGE models such as TransE~\cite{Bordes:TransE} and its variants~\cite{RotatE, QuatE, HAKE} learn the embeddings of entities and relations to score the plausibility of triples for link prediction. 
\begin{figure}
  \centering
  \includegraphics[width=0.47\textwidth]{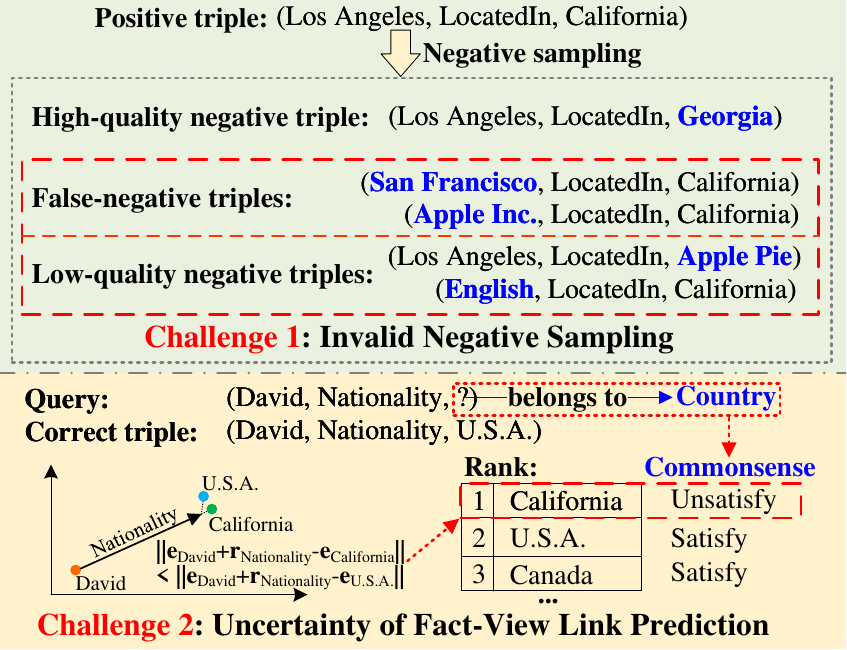}
  \caption{Two examples exhibit the challenges that needed to be addressed. \textbf{Challenge 1}: Given a positive triple, some generated negative triples are false-negative or low-quality. \textbf{Challenge 2}: For link prediction, the entity $California$ ranks higher than the correct entity $U.S.A.$ due to the uncertainty of KG embeddings, but the correct answer entity should belong to the concept $Country$ in the view of commonsense.}
  \label{fig1}
\end{figure}

Among all the existing KGC models, KGE approaches achieve higher efficiency and better performance. Specifically, the KGE-based KGC pipeline can be divided into two stages: learning knowledge graph (KG) embeddings at the training and link prediction at the inference. Learning KG embeddings relies on a basic procedure of negative sampling~\cite{nonesampling}. Link prediction aims to infer the missing entity or relation in a triple via ranking the candidate triples' scores in virtue of the learned KG embeddings.

However, the two separate stages both have drawbacks: (1) \textbf{Invalid negative sampling}: all the previous NS~\cite{Wang:TransH, kbgan, RotatE, zhang2019nscaching, typeconstraint} cannot avoid sampling the false-negative triples and low-quality negative triples, simultaneously. For instance, given the positive triple $(Los\ Angeles,$ $LocatedIn, California)$ as shown in Figure \ref{fig1}, the existing NS strategies might sample the corrupted triples such as $(San\ Francisco, LocatedIn, California)$, which is actually a missing correct triple namely false-negative triple. On the other hand, the quality of some generated negative triples such as $(San\ Francisco, LocatedIn, Apple\ Pie)$ is too poor to make little sense for training the KGE models. (2) \textbf{Uncertainty of fact-view link prediction}: performing link prediction solely based on facts in a data-driven fashion suffers from uncertainty due to the deviation of KG embeddings compared to the symbolic representations, limiting the accuracy of KGC. Take the tail entity prediction $(David, Nationality, ?)$ in Figure \ref{fig1} as an instance. The correct tail entity should belong to the concept $Country$ in the view of commonsense. Whereas the entity $California$ that is inconsistent with commonsense even ranks highest via scoring the candidate triples with KG embeddings.

Last but not least, although some KGE approaches exploit external information, including entity types~\cite{Xie:TKRL}, textual descriptions~\cite{DKRL} and images of entities~\cite{Xie:Image}. Such auxiliary information is hard to access and enhances the single representation of entities rather than providing the semantics of commonsense. However, the valuable commonsense is always acquired by the expensive hand annotation~\cite{commonsense-reasoning}, so its high cost leads to relatively low coverage. Besides, the existing large-scale commonsense KGs such as ConceptNet~\cite{conceptnet} only contain the concepts without the links to the corresponding entities, causing them unavailable to the KGC task.

%Therefore, to improve both the learning of KG embeddings and the link prediction from a generalized perspective that could be applicable to any KGE model, three main challenges should be taken into account: (1) How to address the false-negative and the low-quality negative triples, simultaneously. (2) How to make advantage of commonsense to inject extra semantic constraint for link prediction? (3) How to obtain the appropriate commonsense knowledge from KGs efficiently?

To address the above challenges, we propose a novel and scalable \underline{\textbf{C}}ommonsense-\underline{\textbf{A}}ware \underline{\textbf{K}}nowledge \underline{\textbf{E}}mbedding (\textbf{CAKE}) framework to improve the NS in the training of KGE and boost the performance of KGC \textbf{benefited from the self-supervision of commonsense}. In specific, we attempt to automatically construct explicit commonsense via an instance abstraction technique from KGs. Then, contrary to random sampling, we \textbf{purposefully} generate the high-quality negative triples by taking advantage of the commonsense together with the characteristics of complex relations. Furthermore, a multi-view link prediction is conducted to determine the entity candidates that belong to the correct concepts in the commonsense view and predict the answer entities with the learned KG embeddings from the perspective of fact. In summary, the contributions of our work are three-fold:

\begin{itemize} % 在方法部分说清training和inference stage.
\item We propose a \textbf{scalable} KGC framework with an \textbf{automatic commonsense generation} mechanism to extract valuable commonsense from factual triples and entity concepts.

\item We develop a \textbf{commonsense-aware negative sampling} strategy for generating valid and high-quality negative triples. Meanwhile, a \textbf{multi-view link prediction} mechanism is proposed to improve the accuracy of KGC.

\item Extensive experiments on four benchmark datasets illustrate \textbf{the effectiveness and the scalability} of our whole framework and each module.  The source code and datasets of this paper can be obtained from \url{https://github.com/ngl567/CAKE}.
\end{itemize}

%Most of the KGC (KGC) models merely regard KGs as simple graph-structured data and leverage the instance-level knowledge to predict missing links between entities in data-driven fashion, ignoring the valuable commonsense to complete KGs in the semantic level. Although some approaches employ ontologies of KGs to inject abstract knowledge into learning KG embeddings, the semantics in an ontology always indicates the hierarchical types of entities and is not elaborate enough for accurately representing the commonsense for KGC. Besides, the previous NS strategies without any commonsense suffer from false-negative issue, which degrades the quality of negative instances and further limits the KGC's performance. 

%Although some approaches employ ontologies of KGs to inject abstract knowledge into learning KG embeddings, the semantics in an ontology always indicates the hierarchical types of entities and is not elaborate enough for accurately representing the commonsense for KGC.

\section{Related Work}
\subsection{KGC Models}
The existing KGC models can be classified into three main categories: (1) Rule learning-based algorithms such as AMIE+~\cite{Galarrage:AMIE}, DRUM~\cite{DRUM} and AnyBurl~\cite{AnyBurl} automatically mine logic rules from KGs and apply these rules for inductive link prediction. However, these models are inefficient due to the time-consuming rule searching and evaluation. (2) Path-based models search paths linking head and tail entities, including path ranking approaches~\cite{Lao:PRA, AttnPRA} and reinforcement learning-based models~\cite{DeepPath, Lin:reward-shaping}. Whereas, multi-hop path-based models also spend much time in path searching. (3) KG embedding (KGE) models such as TransE~\cite{Bordes:TransE}, RESCAL~\cite{RESCAL}, ComplEx~\cite{Trouillon:ComplEx}, RotatE~\cite{RotatE} and HAKE~\cite{HAKE} learn the embeddings of entities and relations to score the plausibility of triples for predicting the missing triples efficiently. KGE approaches achieve higher efficiency and better performance on KGC compared with the others. However, the natural uncertainty of embeddings limits the precision of KGC relying solely on facts. More specifically, the KGE models generally need a primary negative sampling (NS) procedure to randomly or purposely sample some triples that are not observed in the KG as negative triples for training~\cite{nonesampling}.

\subsection{Negative Sampling of KGE}
Following the local closed-world assumption \cite{closedworldassumption}, the existing NS techniques for KGE can be classified into five categories: (1) Randomly and uniformly sampling: the majority of the KGE models generate negative triples via randomly replacing an entity or relation in a positive triple from a uniform distribution~\cite{Wang:TransH}. (2) Adversarial-based sampling: KBGAN~\cite{kbgan} integrates the KGE model with softmax probabilities to select the high-quality negative triples in an adversarial training framework. Self-adversarial sampling~\cite{RotatE} performs similar to KBGAN, but it utilizes a self-scoring function without a generator. (3) Domain-based sampling: domain-based NS~\cite{DomainSampling} and type-constrained NS~\cite{typeconstraint} both leverage domain or type constraints on sampling the corrupted entities that belong to the correct domain. (4) Efficient sampling: NSCaching~\cite{zhang2019nscaching} employs cache containing candidates of negative triples to improve the efficiency of sampling. (5) None-sampling: NS-KGE~\cite{nonesampling} eliminates the NS procedure by converting loss functions of KGE into a unified square loss.

However, all the previous NS algorithms cannot address the issue of false-negative triples since these NS techniques, except for none sampling, would attempt to sample the corrupted triples with higher probability while they might be correct and just missing in the KG. Domain-based NS relies heavily on the constraint of the single type rather than the commonsense, limiting the diversity of negative triples. KBGAN introduces generative adversarial networks (GAN) in the NS framework, making the original model more complex and hard to train. None sampling eliminates the negative triples and has to convert each original KGE model into square loss, which weakens the performance of KGE models. These drawbacks of the NS strategies degrade the training of KGE and further limit the performance of KGC.
 
 \subsection{Commonsense Knowledge Graph}
 Different from the factual triples, commonsense could inject rich abstract knowledge into KGs. However, the valuable commonsense is hard to access due to the costly hand annotation. In recent years, many researches attempt to construct general commonsense graphs such as ConceptNet~\cite{conceptnet}, Microsoft Concept Graph~\cite{MicrosoftCG} and ATOMIC~\cite{atomic}. However, these commonsense graphs only contain the concepts without the links to the corresponding entities, causing them inapplicable to the KGC task. On the other hand, although some KGE models such as JOIE~\cite{JOIE} employ the ontology built-in most of the KGs, i.e., NELL~\cite{Mitchell:nell} and DBpedia~\cite{Lehmann:dbpedia}, the relations in ontology such as $isA$, $partOf$ and $relatedTo$ mainly represent the type hierarchy but not the explicit commonsense. Such relations are useless for KGC because there are few overlaps between the ontological and the factual relations.

\begin{figure*}
  \centering
  \includegraphics[width=1\textwidth]{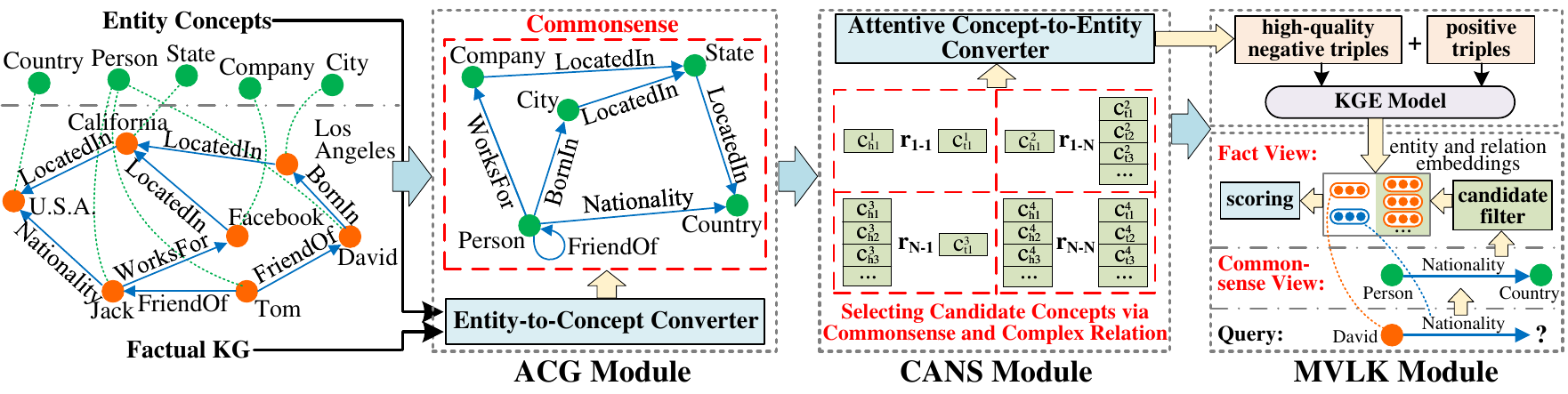}
  \caption{An overview of the CAKE framework. The orange dotes indicate the entities. The green dotes represent the entity concepts. In the CANS module, $r_{1-1}$, $r_{1-N}$, $r_{N-1}$ and $r_{N-N}$ denote the diverse complex relations of 1-1, 1-N, N-1 and N-N, respectively. $c^{j}_{hi}$ and $c^{j}_{ti}$ indicate the $i$-th head concept and tail concept that are selected by the commonsense and the characteristics of complex relations specific to the $j$-th relation.}
  \label{fig2}
\end{figure*}

\section{Methodology}
\label{sec3}
In this section, we introduce our novel and scalable CAKE framework. As shown in Figure \ref{fig2}, the entire pipeline consists of three developed modules: the automatic commonsense generation (ACG) module, the commonsense-aware negative sampling (CANS) module and the multi-view link prediction (MVLP) module. Firstly, the ACG module extracts the commonsense from the factual triples with the entity concepts via an instance abstraction mechanism (\cref{sec3.2}). Then, the CANS module employs the generated commonsense to produce the high-quality negative triples, which takes the characteristics of complex relations into account (\cref{sec3.3}). Afterwards, our approach feeds the positive and the weighted negative triples into the KGE model for learning entity and relation embeddings (\cref{sec3.4}). Finally, the MVLP module conducts link prediction in a coarse-to-fine fashion by filtering the candidates in the view of commonsense and predicting the answer entities with KG embeddings from the candidates in the view of fact (\cref{sec3.5}).

\subsection{Notations and Problem Formalization}
\label{sec3.1}

\noindent \textbf{Commonsense.}
Commonsense has gained widespread attraction from its successful use in understanding high-level semantics, which is generally represented as the concepts with their ontological relations in some well-known commonsense graphs such as ConceptNet~\cite{conceptnet} and Microsoft Concept Graph~\cite{MicrosoftCG}. Notably, we extend the commonsense in two forms: the individual form $\mathcal{C}_1$ and the set form $\mathcal{C}_2$. Both $\mathcal{C}_1$ and $\mathcal{C}_2$ are the sets of triples while each triple in $\mathcal{C}_1$ is constituted of a head entity's concept $c_h$ and a tail entity's concept $c_t$ associated with their instance-level relation $r$, which can be written as follows:
\begin{equation}
    \mathcal{C}_1 = \{ (c_h, r, c_t) \} \label{eq1}
\end{equation}

On the contrary, each triple in $\mathcal{C}_2$ consists of a relation $r$ linking the corresponding head concept set $C_h$ and tail concept set $C_t$, which is shown as:
\begin{equation}
    \mathcal{C}_2 = \{ (C_h, r, C_t) \} \label{eq2}
\end{equation}

The detailed description of commonsense generation is introduced in section \ref{sec3.2}.

\vspace{0.2cm}
\noindent \textbf{KGE Score Function.}
We could leverage any KGE model to learn the entity and relation embeddings owing to our scalable framework independent of the KGE model. Thus, we define a uniform symbol $E(h, r, t)$ to represent the score function of any KEG model for evaluating the plausibility of a triple $(h ,r, t)$. More specifically, the three most typical score function patterns are given as follows: 

(1) The \textbf{translation}-based score function, such as TransE~\cite{Bordes:TransE}:
\begin{equation}
    E(h, r, t) = \Vert \textbf{h} + \textbf{r} - \textbf{t} \Vert \label{eq3}
\end{equation}
where $\textbf{h}$, $\textbf{r}$ and $\textbf{t}$ denote the embeddings of head entity $h$, relation $r$ and tail entity $t$, respectively.

(2) The \textbf{rotation}-based score function, such as RotatE~\cite{RotatE}:
\begin{equation}
    E(h, r, t) = \Vert \textbf{h} \circ \textbf{r} - \textbf{t} \Vert \label{eq4}
\end{equation}
where $\circ$ indicates the hardmard product.

(3) The \textbf{tensor decomposition}-based score function, such as DistMult~\cite{Distmult}:
\begin{equation}
    E(h, r, t) = \textbf{h}^\top diag(\textbf{M}_r) \textbf{t} \label{eq5}
\end{equation}
where $diag(\textbf{M}_r)$ represents the diagonal matrix of the relation $r$.

\vspace{0.2cm}

\noindent \textbf{Link Prediction.}
Following most of the previous KGC models, we regard link prediction as an entity prediction task. Given a triple query with an entity missing $(h, r, ?)$ or $(?, r, t)$, link prediction takes every entity as a candidate. It calculates the score of each candidate triple by employing the learned KG embeddings and the score function. Then, we rank the candidate entities in light of their scores and output the top n entities as results.

\subsection{Automatic Commonsense Generation}
\label{sec3.2}

In terms of the representation of commonsense defined in section~\ref{sec3.1}, our approach could theoretically generate commonsense from any KG automatically as long as there exist some concepts linked to the entities in the KG. Specifically, we develop an entity-to-concept converter to replace the entities in each factual triple with corresponding concepts. Meanwhile, the relations in commonsense entail the instance-level relations in factual KGs. Take an instance in Figure~\ref{fig2}, the factual triple $(David, Nationality, U.S.A.)$ can be transformed to a concept-level triple $(Person,$ $Nationality, Country)$. Particularly, the commonsense in the individual form $\mathcal{C}_1$ is achieved by wiping out the reduplicated concept-level triples. Afterwards, we merge the concept-level triples that contain the same relation into a set to construct the commonsense in the set form $\mathcal{C}_2$.

\begin{figure}
  \centering
  \includegraphics[width=0.47\textwidth]{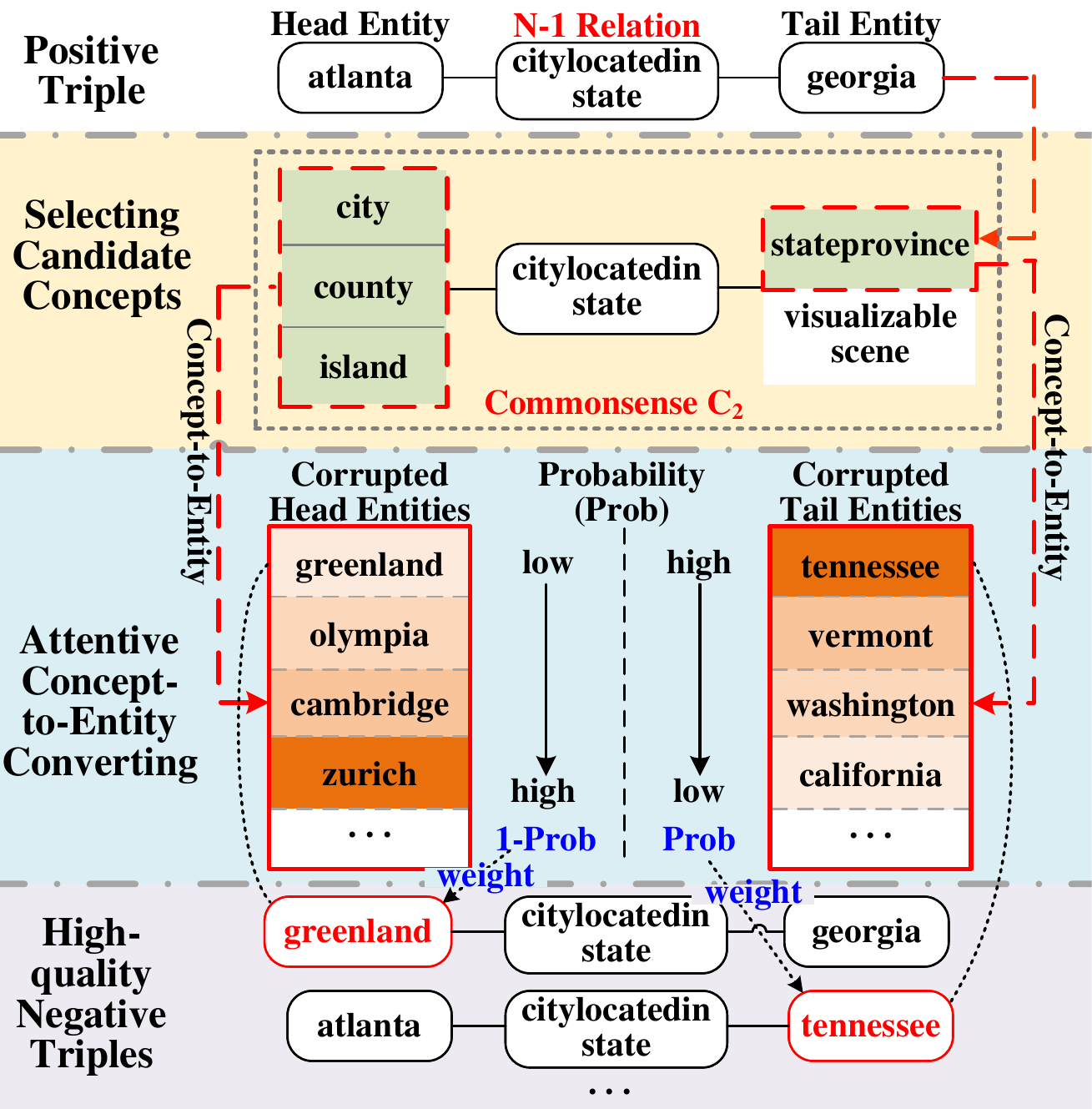}
  \caption{An example of generating the high-quality negative triples containing an N-1 relation by our designed CANS module on NELL-995.}
  \label{fig3}
\end{figure}

\subsection{Commonsense-Aware Negative Sampling}
\label{sec3.3}

Intuitively, the negative triples satisfying commonsense are more challenging to distinguish from positive triples, contributing to more effective training signals. Therefore, we try to sample the negative triples that conform to the commonsense.

To reduce the false-negative triples, we exploit the characteristics of complex relations, namely 1-1, 1-N, N-1, and N-N defined in TransH~\cite{Wang:TransH} for negative sampling, where $\textbf{1}$ implies that the entity is unique when given the relation and another entity, on the contrary, $\textbf{N}$ denotes that there might be multiple entities in this case (non-unique entity). Based on this observation, two specific sampling strategies are proposed: (1) \textbf{uniqueness sampling}: in terms of corrupting a unique entity such as the tail entity of the N-1 relation, the corrupted triples except for the original positive one are definitely actual negative triples. Furthermore, the corrupted entities that share at least one concept with the correct entity are regarded as high-quality negative triples, contributing to a more consistent training signal. (2) \textbf{None-unique sampling}: for corrupting a non-unique entity such as a head entity linked by the N-1 relation, the entities belonging to the same concept(s) with the correct entity are more likely to be false-negative due to the non-uniqueness of the head entity. Thus, the weights of these negative triples being false-negative should be as low as possible in training. Meanwhile, we try to sample the triples conforming to the commonsense $\mathcal{C}_2$ for high quality.

For a better understanding, an example of generating high-quality negative triples with an N-1 relation is shown in Figure \ref{fig3}. The whole NS procedure can be divided into two steps. \textbf{Step 1}: selecting the candidate concepts with commonsense $\mathcal{C}_2$. The candidate head concepts $city$, $county$ and $island$ are determined according to commonsense $\mathcal{C}_2$ and non-unique sampling. Besides, based on the uniqueness sampling strategy, the candidate tail concept is selected as the same concept $stateprovince$ as that of $Georgia$. \textbf{Step 2}: attentive concept-to-entity converting. To reduce false-negative while ensuring the high quality of the negative triples, the corrupted entities belonging to the candidate concepts are sampled from the following distribution:
\begin{align}
    w(h'_j, r, t) &= 1 - p((h'_j, r, t)|\{(h_i, r_i, t_i)\}) \nonumber \\
                 &= 1 - \frac{\exp\alpha E(h'_j, r, t)}{\sum_i{\exp\alpha E(h'_i, r, t)}} \label{eq6} \\
    w(h, r, t'_j) &= p((h, r, t'_j)|\{(h_i, r_i, t_i)\}) \nonumber \\
                   &= \frac{\exp\alpha E(h, r, t'_j)}{\sum_i{\exp\alpha E(h, r, t'_i)}} \label{eq7}
\end{align}
where $h'_i$ and $t'_i$ are the corrupted head and tail entities obtained by non-unique sampling and uniqueness sampling. $w$ and $p$ denote the weight and the probability of the negative triple, respectively. $\alpha$ is the temperature of sampling motivated by the self-adversarial sampling~\cite{RotatE}. Remarkably, considering that a triple with a higher probability is more likely to be a positive one, the weight of a negative triple containing the corrupted head entity such is defined as Eq.~\ref{eq6} to prevent the issue of false-negative. Besides, the negative triples containing the corrupted tail entities with higher probability are endowed with higher-quality weight since there is no false-negative issue. Thus, both the corrupted head entity $greenland$ and the corrupted tail entity $tennessee$ with the high weights are selected to generate high-quality negative triples.

\subsection{Traning the KGE Model}
\label{sec3.4}
Based on the negative triples obtained by CANS, we train the KGE model to learn the entity and relation embeddings for enlarging the gap between the scores of the positive and high-quality negative triples. In this work, we employ the following loss function as our optimization objective:
\iffalse
\begin{equation}\small
\begin{split}
    L = \sum_{i}^{n}{[\gamma+E(h,r,t)-w(h'_i, r, t)E(h'_i, r, t)]_{+}} \\
    +[\gamma+E(h,r,t)-w(h, r, t'_i)E(h, r, t'_i)]_{+} \label{eq8}
\end{split}
\end{equation}
or
\fi
\begin{equation}\small
\begin{split}
    L = &-log\sigma(\gamma-E(h, r, t)) \\
    &- \sum_{i}^{n}{0.5[w(h'_i, r, t) log\sigma(E(h'_i, r, t)-\gamma)} \label{eq9} \\
    &\ \ \ \ \ \ \ \ \ \ \ \ \ \ \ \ \ +w(h, r, t'_i) log\sigma(E(h, r, t'_i)-\gamma)] 
\end{split}
\end{equation}
in which $\gamma$ is the margin. $[x]_+$ indicates the larger value between 0 and $x$. $\sigma$ is the sigmoid function.

\subsection{Multi-View Link Prediction}
\label{sec3.5}
Benefiting from the same relations among commonsense and facts, commonsense could directly provide a definite range for link prediction results. Hence we develop a novel multi-view link prediction (MVLK) mechanism in a coarse-to-fine paradigm to facilitate more likely predicted results. Firstly, at the coarse prediction stage, we pick out the candidate entities in the view of commonsense. Specifically, take a query $(h , r, ?)$ for an example, commonsense $\mathcal{C}_1$ is employed for filtering the reasonable concepts of the tail entity. The candidate concept set of tail entity is defined as
\begin{equation}
    \textbf{C}_{1t} = \{c_{ti} | (c_{hi}, r, c_{ti})\in \mathcal{C}_1\} \label{eq10}
\end{equation}
where $c_{hi}$ is the i-th concept of $h$, and $c_{ti}$ denotes the tail concept in the commonsense $(c_{hi}, r, c_{ti})$. Then, the entities belonging to the concept set $\textbf{C}_{1t}$ can be determined as the candidate entities since they satisfy commonsense and are more likely to be the correct tail entities from the perspective of commonsense compared with other entities.

Then, at the fine prediction stage, we score each candidate entity $e_i$ derived from the coarse prediction stage in the view of fact as following
\begin{equation}
    score(e_i) = E(h, r, e_i)
\end{equation}
where $E(h, r, e_i)$ denotes the score function employed for training the KGE model. Subsequently, the prediction results will rank the scores of candidate entities in ascending order and output the entities with higher ranks.

\section{Experiments and Results}
In this section, we perform extensive experiments of KGC on four widely-used KG datasets containing concepts. We firstly describe datasets, baseline models, implementation details and evaluation protocol. Then, the effectiveness of our proposed framework CAKE and each module is demonstrated by compared with several baselines. Furthermore, we conduct further experiments, including the ablation study and the case study.

\begin{table}\small
 \centering
 \renewcommand\tabcolsep{2.5pt}
 \renewcommand{\arraystretch}{1.0}
 \begin{tabular}{c|ccc|ccc}
 \toprule
Dataset		& \#Rel	& \#Ent	& \#Con	 & \#Train	& \#Valid	& \#Test \\
 \midrule
 FB15K		& 1,345		    & 14,951    & 89        & 483,142	& 50,000	& 59,071 \\
 FB15K237   & 237           & 14,505    & 89        & 272,115   & 17,535    & 20,466 \\
 NELL-995   & 200           & 75,492    & 270       & 123,370   & 15,000    & 15,838 \\
 DBpedia-242 & 298          & 99,744    & 242       & 592,654   & 35,851    & 30,000 \\
 \bottomrule
 \end{tabular}
 \caption{Statistics of the experimental datasets. \#Rel, \#Ent, \#Con represent the number of relations, entities and concepts of each dataset, respectively.}
 \label{table1}
 \end{table}

\begin{table*}[!t]\small
%\setlength{\abovecaptionskip}{-0.005cm}
%% increase table row spacing, adjust to taste
%\renewcommand{\arraystretch}{0.6}
\centering
\renewcommand\tabcolsep{6pt}
\begin{tabular}{l|ccccc|ccccc}
\toprule
\multirow{2}*{Models} & \multicolumn{5}{c|}{FB15K} & \multicolumn{5}{c}{FB15K237} \\
	& MR	& MRR	& Hits@10	& Hits@3  & Hits@1	& MR	& MRR	& Hits@10	& Hits@3  & Hits@1\\
\midrule
TransE         & 35   & 0.626     & 0.838     & 0.723     & 0.496       & 195     & 0.268     & 0.454     & 0.298     & 0.176 \\
TransE+\textbf{CANS}	        & 34   & 0.671    & 0.864    & \textbf{0.761}	& 0.552     & \textbf{175} & 0.298  & 0.490      & 0.333   & 0.203  \\
TransE+\textbf{MVLP}	        & 35   & 0.636    & 0.839    & 0.725	&0.513     & 181 & 0.290  & 0.476      & 0.323   & 0.186  \\
TransE+\textbf{CAKE}	        & \textbf{33}   & \textbf{0.672}    & \textbf{0.865}    & \textbf{0.761}	& \textbf{0.555}     & \textbf{175} & \textbf{0.301}  & \textbf{0.493}      & \textbf{0.335}   & \textbf{0.206}  \\
\midrule
\iffalse
DistMult            & 46   & 0.499     & 0.734     & 0.576     & 0.369       & 230     & 0.307     & 0.507     & 0.342     & 0.209 \\
DistMult+\textbf{CANS}	        & \textbf{20}   & \textbf{0.854}    & \textbf{0.933}    & \textbf{0.885}	& \textbf{0.810}     & \textbf{118} & \textbf{0.555}  & \textbf{0.707}      & \textbf{0.590}   & \textbf{0.479}  \\
DistMult+\textbf{MVLP}	        & \textbf{20}   & 0.528    & 0.747    & 0.610	& 0.402     & \textbf{118} & \textbf{0.555}  & \textbf{0.707}      & \textbf{0.590}   & \textbf{0.479}  \\
DistMult+\textbf{CAKE}	        & \textbf{20}   & \textbf{0.854}    & \textbf{0.933}    & \textbf{0.885}	& \textbf{0.810}     & \textbf{118} & \textbf{0.555}  & \textbf{0.707}      & \textbf{0.590}   & \textbf{0.479}  \\
\midrule
ComplEx    & 41   & 0.556     & 0.786     & 0.641     & 0.425       & 197     & 0.265     & 0.434     & 0.291     & 0.182 \\
ComplEx+\textbf{CANS}	        & \textbf{20}   & \textbf{0.854}    & \textbf{0.933}    & \textbf{0.885}	& \textbf{0.810}     & \textbf{118} & \textbf{0.555}  & \textbf{0.707}      & \textbf{0.590}   & \textbf{0.479}  \\
ComplEx+\textbf{MVLP}	        & \textbf{20}   & \textbf{0.854}    & \textbf{0.933}    & \textbf{0.885}	& \textbf{0.810}     & \textbf{118} & 0.304  & 0.497      & 0.338   & 0.208  \\
ComplEx+\textbf{CAKE}	        & \textbf{20}   & \textbf{0.854}    & \textbf{0.933}    & \textbf{0.885}	& \textbf{0.810}     & \textbf{118} & \textbf{0.555}  & \textbf{0.707}      & \textbf{0.590}   & \textbf{0.479}  \\
\midrule
\fi
RotatE    & 35   & 0.657     & 0.850     & 0.746     & 0.537      & 204     & 0.269     & 0.452     & 0.298     & 0.179 \\
RotatE+\textbf{CANS}	        & 33   & 0.702    & 0.877       & 0.790    & 0.588     & 182   & 0.296    & 0.486    & 0.329	& 0.202   \\
RotatE+\textbf{MVLP}	        & 34   & 0.688    & 0.860       & 0.768    & 0.579	    & 188 & 0.308  & 0.493      & 0.340   & 0.217  \\
RotatE+\textbf{CAKE}	        & \textbf{31}   & \textbf{0.705}    & \textbf{0.878}    & \textbf{0.792}	& \textbf{0.593}     & \textbf{181} & \textbf{0.318}  & \textbf{0.511}      & \textbf{0.354}   & \textbf{0.223}  \\
\midrule
HAKE                    & 34    & 0.690     & 0.872     & 0.780     & 0.574      & 176     & 0.306     & 0.486     & 0.337     & 0.216 \\
HAKE+\textbf{CANS}	        & 37   & 0.723    & 0.882    & 0.808	& 0.616     & 174 & 0.315  & 0.501      & 0.344   & 0.221  \\
HAKE+\textbf{MVLP}	        & 32   & 0.729    & 0.890    & 0.817	& 0.622     & 172 & 0.320  & 0.508      & 0.352   & 0.227  \\
HAKE+\textbf{CAKE}	        & \textbf{30}   & \textbf{0.741}    & \textbf{0.896}    & \textbf{0.825}	& \textbf{0.646}     & \textbf{170} & \textbf{0.321}  & \textbf{0.515}      & \textbf{0.355}   & \textbf{0.226}  \\
\bottomrule
\toprule
\multirow{2}*{Models} & \multicolumn{5}{c|}{DBpedia-242} & \multicolumn{5}{c}{NELL-995} \\
	& MR	& MRR	& Hits@10	& Hits@3  & Hits@1	& MR	& MRR	& Hits@10	& Hits@3  & Hits@1\\
\midrule
TransE         & 2733   & 0.242     & 0.468    & 0.344     & 0.100    & 1081      & 0.429     & 0.557     & 0.477      & 0.354 \\
TransE+\textbf{CANS}	        & 1889   & 0.287    & 0.575    & 0.427	& 0.103      & 1022   & 0.433    & 0.591    & 0.495	& 0.336   \\
TransE+\textbf{MVLP}	        & \textbf{881}   & 0.322    & 0.585    & 0.450	& 0.152      & 336   & 0.509    & 0.617    & 0.547	& 0.444  \\
TransE+\textbf{CAKE}	        & \textbf{881}   & \textbf{0.330}    & \textbf{0.595}    & \textbf{0.458}	& \textbf{0.160}  & \textbf{317}   & \textbf{0.533}    & \textbf{0.650}    & \textbf{0.578}	& \textbf{0.461} \\
\midrule
\iffalse
DistMult            & 12289      & 0.195     & 0.337     & 0.229     & 0.122       & 7241     & 0.165     & 0.250     & 0.172     & 0.122 \\
DistMult+\textbf{CANS}	        & \textbf{20}   & \textbf{0.854}    & \textbf{0.933}    & \textbf{0.885}	& \textbf{0.810}     & \textbf{118} & \textbf{0.555}  & \textbf{0.707}      & \textbf{0.590}   & \textbf{0.479}  \\
DistMult+\textbf{MVLP}	        & \textbf{20}   & \textbf{0.854}    & \textbf{0.933}    & \textbf{0.885}	& \textbf{0.810}     & \textbf{118} & \textbf{0.555}  & \textbf{0.707}      & \textbf{0.590}   & \textbf{0.479}  \\
DistMult+\textbf{CAKE}	        & \textbf{20}   & \textbf{0.854}    & \textbf{0.933}    & \textbf{0.885}	& \textbf{0.810}     & 1109 & 0.438  & 0.524      & 0.470   & 0.383  \\
\midrule
ComplEx    & 2750   & 0.141     & 0.241    & 0.155     & 0.088    & -     & 0.169     & 0.298     & 0.185      & 0.106 \\
ComplEx+\textbf{CANS}	        & \textbf{20}   & \textbf{0.854}    & \textbf{0.933}    & \textbf{0.885}	& \textbf{0.810}     & \textbf{118} & \textbf{0.555}  & \textbf{0.707}      & \textbf{0.590}   & \textbf{0.479}  \\
ComplEx+\textbf{MVLP}	        & \textbf{20}   & \textbf{0.854}    & \textbf{0.933}    & \textbf{0.885}	& \textbf{0.810}     & \textbf{118} & \textbf{0.555}  & \textbf{0.707}      & \textbf{0.590}   & \textbf{0.479}  \\
ComplEx+\textbf{CAKE}	        & \textbf{20}   & \textbf{0.854}    & \textbf{0.933}    & \textbf{0.885}	& \textbf{0.810}     & \textbf{118} & \textbf{0.555}  & \textbf{0.707}      & \textbf{0.590}   & \textbf{0.479}  \\
\midrule
\fi
RotatE                & 1950   & 0.374     & 0.582     & 0.457     & 0.249        & 2077       & 0.460     & 0.553    & 0.493  & 0.403 \\
RotatE+\textbf{CANS}	        & 1063   & 0.407    & 0.593    & 0.476	& 0.300      & 1097   & 0.531    & 0.644    & 0.573	& 0.461   \\
RotatE+\textbf{MVLP}	       & \textbf{983}   & 0.393    & 0.594    & 0.474	& 0.273   & 356   & 0.519    & 0.628    & 0.564	& 0.447 \\
RotatE+\textbf{CAKE}	        & 1027    & \textbf{0.423}    & \textbf{0.603}    & \textbf{0.486}	& \textbf{0.320}      & \textbf{329}   & \textbf{0.546}    & \textbf{0.660}    & \textbf{0.592}	& \textbf{0.474}   \\
\midrule
HAKE                    & 1757     & 0.408     & 0.579     & 0.463    & 0.312     & 1157    & 0.502     & 0.610      & 0.538   & 0.437 \\
HAKE+\textbf{CANS}	        & 1147   & 0.427    & 0.587    & 0.472	& 0.341     & 2011 & 0.520  & 0.640      & 0.556   & 0.451  \\
HAKE+\textbf{MVLP}	        & 1083   & 0.411    & 0.580    & 0.463	& 0.319     & 478 & 0.510  & 0.614      & 0.551   & 0.444  \\
HAKE+\textbf{CAKE}	        & \textbf{931}   & \textbf{0.437}    & \textbf{0.593}    & \textbf{0.481}	& \textbf{0.353}     & \textbf{433} & \textbf{0.543}  & \textbf{0.655}      & \textbf{0.583}   & \textbf{0.477}  \\
\bottomrule
\end{tabular}
\caption{Link prediction results on four datasets. \textbf{Bold} numbers are the best results for each type of model.}
\label{table2}
\end{table*}

\subsection{Experiment Settings}
\noindent \textbf{Datasets.}
Four real-world datasets containing ontological concepts are utilized for experiments, including FB15K~\cite{Bordes:TransE}, FB15K237~\cite{FB15k237}, NELL-995~\cite{DeepPath} and DBpedia-242. Particularly, DBpedia-242 is extracted from DBpedia~\cite{Lehmann:dbpedia} which contains totally 242 concepts. The statistics of the datasets are summarized in Table~\ref{table1}. Notably, the entities in FB15K and FB15K237 always belong to more than one concept while each entity in NELL-995 and DBpedia-242 has only one concept.

\noindent \textbf{Baselines.}
We compare our CAKE model with three state-of-the-art KGE models, including TransE~\cite{Bordes:TransE}, RotatE~\cite{RotatE} and HAKE~\cite{HAKE}. Meanwhile, these baselines are also the basic models integrated with our framework. It is unnecessary to use many baselines since the focus of this work is to observe the impact of applying our CAKE framework to original KGE models instead of defeating all the SOTA models. We provide the results of baselines by running their source codes$\footnote{The codes of TransE and RotatE: \url{https://github.com/DeepGraphLearning/KnowledgeGraphEmbedding}. The code of HAKE: \url{https://github.com/MIRALab-USTC/KGE-HAKE}.}$ with the suggested parameters. Note that all the existing type-based and ontology-based models are not chosen as baselines since they are specific to a few KGs and cannot work on most of the datasets in our experiment.

\noindent \textbf{Implementation Details.}
Each complex relation is labelled in the same way as in TransH~\cite{Wang:TransH}. We use Adam optimizer for the training and tune the hyper-parameters of our model by grid search on the validation sets. Specifically, the embedding size and the batch size are the same as those of each basic model for a fair comparison. The negative sampling size is set as 2 for all the models. The learning rate is chosen from 0.0001 to 0.01. The margin is tuned in \{9, 12, 18, 24, 30\}. The sampling temperature is adjusted in \{0.5, 1.0\}. All the experiments are conducted in Pytorch and on GeForce GTX 2080Ti GPUs.

\noindent \textbf{Evaluation Protocol.}
Following the procedure of MVLP in Section \ref{sec3.5}, we can obtain the rank of the correct entity for each test example. Then, the performance of link prediction is evaluated by three commonly-used metrics: mean rank (MR), mean reciprocal rank (MRR), and proportion of the correct entities ranked in the top n (Hits@N). All the metrics are in the filtered setting by wiping out the candidate triples already exist in the datasets.

\subsection{Experimental Results}
Table \ref{table2} exhibits the evaluation results of link prediction on the four datasets. We can observe that both CANS and MVLP modules effectively improve the performance of each basic model on each dataset. Moreover, the entire CAKE framework further facilitates more performance gains than each separate module and outperforms all the baselines consistently and significantly. Compared with the performance average of the three baseline models, our CAKE model improves MRR by \textbf{7.2}\%, \textbf{11.5}\%, \textbf{16.2}\% and \textbf{16.7}\% on FB15K, FB15K237, DBpedia-242 and NELL-995. These results demonstrate the superiority and effectiveness of integrating commonsense with the original KGE models.

\begin{table*}[!t]\small
%\setlength{\abovecaptionskip}{-0.005cm}
%% increase table row spacing, adjust to taste
%\renewcommand{\arraystretch}{1.0}
\centering
\renewcommand\tabcolsep{3.5pt}
\begin{tabular}{l|ccccc|ccccc}
\toprule
\multirow{2}*{Models} & \multicolumn{5}{c|}{FB15K} & \multicolumn{5}{c}{FB15K237} \\
	& MR	& MRR	& Hits@10	& Hits@3  & Hits@1	& MR	& MRR	& Hits@10	& Hits@3  & Hits@1\\
\midrule
TransE+Unifo         & 178 & 0.301     & 0.505     & 0.339     & 0.201     & 361    & 0.171    & 0.323     & 0.182     & 0.097 \\
TransE+NoSamp        & 144  & 0.350     & 0.578     & 0.415     & 0.227       & 343      & 0.261    & 0.446     & 0.297     & 0.168 \\
TransE+NSCach      & 209   & 0.292     & 0.560     & 0.375     & 0.144       & 556      & 0.205    & 0.353     & 0.226     & 0.131 \\
TransE+Domain      & 35    & 0.619     & 0.839     & 0.715     & 0.489        & 186    & 0.283     & 0.467     & 0.314     & 0.190 \\
TransE+SAdv         & 35  & 0.626     & 0.838     & 0.723     & 0.496       & 195      & 0.268    & 0.454     & 0.298     & 0.176 \\
\textbf{TransE+CANS (Ours)}        & \textbf{34}   & \textbf{0.671}    & \textbf{0.864}    & \textbf{0.761}	& \textbf{0.552}    & \textbf{175}      & \textbf{0.298}    & \textbf{0.490}     & \textbf{0.333}     & \textbf{0.203} \\
\iffalse
\midrule
ComplEx+Unifo         &      & 0.     & 0.     & 0.       & 0.    & &   & 0.     & 0.     & 0. \\
ComplEx+None       &    & 0.263     & 0.422     & 0.290     & 0.180       &     & 0.237    & 0.385     & 0.257     & 0.164 \\
ComplEx+NSCach     &    & 0.338     & 0.553     & 0.374     & 0.235       &     & 0.207    & 0.353     & 0.230     & 0.133 \\
ComplEx+Domain     &    &      & 0.     & 0.     & 0.       &   & 0.    & 0.     & 0.     & 0. \\
ComplEx+SAdv       &     & 0.556     & 0.786     & 0.641     & 0.425    &      & 0.266    & 0.434     & 0.291     & 0.182 \\
\textbf{ComplEx+CANS (Ours)}       &    & \textbf{0.670}     & \textbf{0.832}     & \textbf{0.727}     & \textbf{0.579}       &     & \textbf{0.278}    & \textbf{0.442}     & \textbf{0.302}     & \textbf{0.197} \\
\midrule
DistMult+Unifo     &  & 0.     & 0.     & 0.     & 0.       &   & 0.    & 0.     & 0.     & 0. \\
DistMult+None      &    & 0.198     & 0.362     & 0.209     & 0.120       &     & 0.149    & 0.262     & 0.153     & 0.092 \\
DistMult+NSCach    &    & 0.488     & 0.728     & 0.559     & 0.364       &     & 0.281    & 0.443     & 0.305     & 0.201 \\
DistMult+Domain         &      & 0.     & 0.     & 0.       & 0.    &       & 0.     & 0.     & 0. \\
DistMult+SAdv       &   & 0.499     & 0.734     & 0.576     & 0.369       &     & 0.307    & 0.507     & 0.342     & 0.209 \\
\textbf{DistMult+CANS (Ours)}       &   & \textbf{0.563}     & \textbf{0.758}     & \textbf{0.642}     & \textbf{0.447}       &     & 0.305    & 0.496     & 0.339     & \textbf{0.210} \\
\fi
\bottomrule
\toprule
\multirow{2}*{Models} & \multicolumn{5}{c|}{DBpedia-242} & \multicolumn{5}{c}{NELL-995} \\
	& MR	& MRR	& Hits@10	& Hits@3  & Hits@1	& MR	& MRR	& Hits@10	& Hits@3  & Hits@1\\
\midrule
TransE+Unifo         & 5750   & 0.124     & 0.262     & 0.183    & 0.033       & 8650    & 0.167    & 0.354     & 0.219     & 0.068 \\
TransE+None        & 2292    & 0.202     & 0.395     & 0.247     & 0.101       & 9172    & 0.176    & 0.297     & 0.210     & 0.106 \\
TransE+NSCach      & 5465   & 0.156     & 0.340     & 0.212     & 0.050       & 13967    & 0.107    & 0.205     & 0.122     & 0.107 \\
TransE+Domain      & 3415   & 0.203     & 0.510     & 0.346     & 0.009      & 1319       & 0.381    & 0.549     & 0.468     & 0.271 \\
TransE+SAdv         & 2733   & 0.242     & 0.468    & 0.344     & 0.100     & 1081      & 0.429     & 0.557     & 0.477      & \textbf{0.354} \\
\textbf{TransE+CANS (Ours)}        & \textbf{1889}   & \textbf{0.287}    & \textbf{0.575}    & \textbf{0.427}	& \textbf{0.103}       & \textbf{1022}   & \textbf{0.433}    & \textbf{0.591}    & \textbf{0.495}	& 0.336 \\
\bottomrule
\end{tabular}
\caption{Comparison results of various NS techniques. Unifo, NoSamp, NSCach, Domain and SAdv denote uniform sampling, none sampling, NSCaching, domain-based NS and self-adversarial NS strategies, respectively.}
\label{table3}
\end{table*}

We compare our CANS module with various types of NS techniques, including uniform sampling~\cite{Bordes:TransE}, none sampling~\cite{nonesampling}, NSCaching~\cite{zhang2019nscaching}, domain-based sampling~\cite{DomainSampling} and self-adversarial sampling~\cite{RotatE}. The comparison results are obtained by combining these NS techniques with the most classical KGE model TransE\cite{Bordes:TransE}. From the results shown in Table \ref{table3}, our CANS module significantly outperforms all the other NS techniques on all the datasets. Specifically, domain-based NS, self-adversarial sampling and our CANS module consistently outperform the others due to evaluating the quality of negative triples. Furthermore, our CANS module performs better than domain-based NS and self-adversarial sampling since CANS could reduce false-negative. These results illustrate the superior ability of our CANS module to generate more high-quality negative triples for enhancing the performance of any KGE model.

\subsection{Ablation Study}

\begin{figure}
  \centering
  \includegraphics[width=0.47\textwidth]{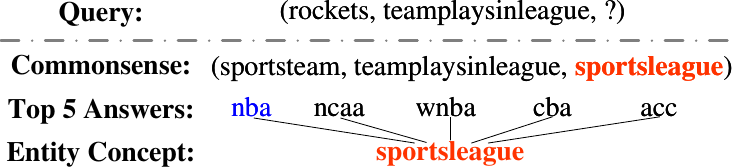}
  \caption{A case study of explainable link prediction with commonsense and entity concept on NELL-995.}
  \label{casestudy}
\end{figure}

We verify the effectiveness of each contribution via integrating the whole framework CAKE and the following ablated models into the basic model HAKE: (1) neglecting the characteristics of complex relations in CANS (-CRNS), (2) removing the commonsense in CANS while retaining the characteristics of complex relations (-CSNS), and (3) omitting the commonsense-view prediction from MVLP (-MVLP). The results in Table \ref{table5} demonstrate that our whole model CAKE performs better than all the ablated models on each dataset. It illustrates that introducing commonsense and the characteristics of complex relations both make sense in the NS process for generating more effective negative triples. Besides, MVLP facilitates link prediction performance benefited from determining the reasonable candidate entities by prior commonsense. In general, each contribution plays a pivotal role in our approach.

\begin{table}\small
\centering
\begin{tabular}{c|ccccc}
\toprule
\multirow{2}*{Models} & \multicolumn{5}{c}{FB15K237} \\
	& MR	& MRR	& Hits@10	& Hits@3  & Hits@1 \\
\midrule
CAKE	                          & \textbf{170}	    & \textbf{0.321}  & \textbf{0.515}      & \textbf{0.355}   & \textbf{0.226}	\\
-CRNS                             & 186		  & 0.318     & 0.507       & 0.352     & 0.223  \\
-CSNS                             & 182		  & 0.317     & 0.509       & 0.351     & 0.222  \\
-MVLP                             & 174       & 0.315     & 0.501       & 0.344     & 0.221  \\
\bottomrule
\toprule
\multirow{2}*{Models} & \multicolumn{5}{c}{NELL-995} \\
	& MR	& MRR	& Hits@10	& Hits@3  & Hits@1 \\
\midrule
CAKE	                      & \textbf{433} & \textbf{0.543}  & \textbf{0.655}      & \textbf{0.583}   & \textbf{0.477}  \\
-CRNS                             & 650		  & 0.519     & 0.627       & 0.564     & 0.453  \\
-CSNS                             & 447		  & 0.529     & 0.647       & 0.567     & 0.463  \\
-MVLP                             & 2011      & 0.520  & 0.640      & 0.556   & 0.451  \\
\bottomrule
\end{tabular}
\caption{Ablation study of integrating each model into the basic model HAKE on FB15K237 and NELL-995.}
\label{table5}
\end{table}

\subsection{Case Study}

We provide the case study of explainable link prediction with commonsense as shown in Figure \ref{casestudy}. Given a query with the tail entity missing $(rockets, teamplaysinleague, ?)$ on NELL-995, our model could output the answer entities and provide the corresponding entity concepts together with the commonsense specific to the query. We can observe that all the top-5 entities including the correct entity $nba$ belong to the concept $sportsleague$ which satisfies the commonsense $(rockets, teamplaysinleague,$ $sportsleague)$. More interestingly, the commonsense and the entity concepts could explain the rationality of the predicted answer entities to enhance the users' credibility of the answers.

\section{Conclusion}
\label{conclusion}

In this paper, we propose a novel and scalable commonsense-aware knowledge embedding framework, which could automatically generate commonsense from KGs with entity concepts for the KGC task. We exploit the generated commonsense to produce effective and high-quality negative triples. On the other hand, we design a multi-view link prediction technique in a coarse-to-fine paradigm to filter the candidate entities in the view of commonsense and output the predicted results from the perspective of fact. The experiments on four datasets demonstrate the effectiveness and the scalability of our proposed framework and each module compared with the state-of-the-art baselines. Furthermore, our framework could explain link prediction results and potentially assemble new KGE models to improve their performance.

\bibliography{anthology}
\bibliographystyle{acl_natbib}

\end{document}